\documentclass[preprint, 5p]{elsarticle}

\def\BibTeX{{\rm B\kern-.05em{\sc i\kern-.025em b}\kern-.08emT\kern-.1667em\lower.7ex\hbox{E}\kern-.125emX}}
    
\makeatletter %
\let\proof\@undefined
\let\endproof\@undefined
\makeatother %

\usepackage{graphicx}
\usepackage{amssymb}
\usepackage{amsthm}
\usepackage{amsmath}
\usepackage{amsfonts}
\usepackage[ruled,noend,linesnumbered]{algorithm2e}
\usepackage{balance}
\usepackage{textcomp}
\usepackage{url}
\usepackage{enumitem} 
\usepackage[labelfont=bf]{caption} 
\usepackage{comment}
\usepackage{todonotes}
\usepackage[normalsize]{subfigure}
\usepackage{booktabs}
\usepackage{multirow}

\specialcomment{ssx}{\begingroup\color{red}Comments: }{\endgroup}
\includecomment{comment} 

\newlength{\figwidths}
\newlength{\figwidthd}
\newlength{\expwidths}
\newlength{\expwidthd}
\theoremstyle{plain}
\newtheorem{definitions}{Definition}

\makeatletter  
\def\@endtheorem{\hfill\ensuremath{\blacksquare}\endtrivlist\@endpefalse } 
\makeatother
\theoremstyle{definition}
\newtheorem{examples}{Example}

\theoremstyle{remark}



\newcommand{\nop}[1]{}

\begin{document}

\title{Time Series Data Imputation: A Survey on Deep Learning Approaches}

\author[1]{Chenguang Fang\corref{cor1}}
\ead{fcg19@mails.tsinghua.edu.cn}

\author[1]{Chen Wang}
\ead{wang_chen@tsinghua.edu.cn}

\address[1]{Tsinghua University, Beijing}

\cortext[cor1]{Corresponding author. Tel.: +(86)18867608173}

\markboth{xxx}%
{xxx}

\begin{abstract}
  Time series are all around in real-world applications. However, unexpected accidents for example broken sensors or missing of the signals will cause missing values in time series, 
  making the data hard to be utilized. It then does harm to the downstream applications such as traditional classification or regression, sequential data integration
  and forecasting tasks, thus raising the demand for data imputation.
  Currently, time series data imputation is a well-studied problem with different categories of methods. 
  However, these works rarely take the temporal relations among the observations and treat the time series as normal structured data, losing the information from the time data.
  In recent, deep learning models have raised great attention. Time series methods based on deep learning have made progress with the usage of models like RNN, 
  since it captures time information from data. In this paper, we mainly focus on time series imputation technique with deep learning methods, which recently made progress in this field.
  We will review and discuss their model architectures, their pros and cons as well as their effects to show the development of the time series imputation methods.
   
\end{abstract}


\maketitle

\section{Introduction}
\label{sect:introduction}
  Time series are vital in real-world applications. However, due to unexpected accidents, for example broken sensors or missing of the signals, 
  missing values are everywhere in time series. In some datasets, the missing rate can reach 90\%, which makes the data hard to be utilized \cite{DBLP:journals/cbm/Garcia-Laencina15}. 
  The missing values significantly do harm to the downstream applications such as traditional classification or regression, sequential data integration \cite{DBLP:conf/icde/LiWZWMH16}
  and forecasting tasks \cite{DBLP:journals/asc/HsiehHY11}, leading to high demand for data imputation. 
  
  Our preliminary study \cite{DBLP:conf/cikm/FangSCG19} shows that imputing the missing values indeed helps significantly the prediction of fuel consumption.
  In the scenarios of fuel consumption prediction, missing values happen due to the errors of sensors. 
  We propose an imputation approach named FuelNet to deal with such errors. The FuelNet generates proper values to impute missing data. With imputed data, the 
  fuel consumption can be reduced by around 45.5\%.

  In current stages, time series data imputation is a well studied problem with different categories of methods including 
  deletion methods, simple imputation methods and learning based methods. However, these works rarely take the temporal relations among the observations and treat the time series as normal structured data, thus losing 
  the information from the time data.

  Fortunately, with the increasing development of deep learning, a large quantity of deep learning methods are researched, among which RNN is one of the typical methods to handle sequence data. 
  The intuition on why deep learning models could advance imputation tasks is that, they are proven to have the ability to mine information hidden in the time series. These characteristics could enable them 
  to impute missing values with such models. 

  Recently, deep learning methods have been applied to multivariable time series imputation and show positive progress in imputing 
  the missing values. In this paper, we mainly survey three papers about time series imputation with deep learning methods \cite{che2018recurrent, DBLP:conf/nips/LuoCZXY18, DBLP:conf/nips/CaoWLZLL18, DBLP:conf/ijcai/Luo0CY19, DBLP:conf/nips/LiuYZZY19}
  among which RNN, GRU and GAN are adopted separately or in combination. We will review these papers about their model structure, the common parts they all adopted and the advantages and disadvantages through comparison.
  
  The remainder of the paper is organized as follows. In the next section, we categorize existing data imputation methods and mainly give an introduction to deep learning imputation methods.
  Section 3 will show the definition of the problems and the symbols. Section 4 will give a detailed discussion of deep learning methods, mainly about their concrete structure, advantages and disadvantages. 
  And finally in Section 5 we summarize the survey and give our conclusions.

\begin{table*}[t]
  \centering
  \caption{Comparison of different methods addressing time series imputation}
  \label{tab:comparison-of-methods}
  \resizebox{\textwidth}{50mm}{ 
  \begin{tabular}{|l|l|l|l|l|}
    \hline
    Methodologies                     & Sample approaches from the literature & Time interval    & Value type               & Time series dimension \\ \hline
    \multirow{2}{*}{Deletion}         & Listwise Deletion \cite{wothke2000longitudinal}                     & regular/irregular  & qualitative & multidimensional        \\ \cline{2-5} 
                                      & Pairwise Deletion \cite{mcknight2007missing}                     & regular/irregular  & qualitative & multidimensional                      \\ \hline
    \multirow{2}{*}{Neighbor Based} & QDORC \cite{DBLP:conf/kdd/SongLZ15}                                        & regular/irregular                & quantitative/qualitative &   multidimensional                     \\ \cline{2-5} 
&  SRKN \cite{icde19} & regular/irregular & quantitative/qualitative &   multidimensional \\ \hline 
    \multirow{2}{*}{Constraint Based} & DERAND \cite{DBLP:journals/pvldb/SongZC015, DBLP:journals/tkde/SongSZCW20}         &  regular/irregular                                      & quantitative/qualitative &  multidimensional                      \\ 
\cline{2-5} & SCREEN \cite{DBLP:conf/sigmod/SongZWY15} & regular/irregular & qualitative & single dimensional \\ \hline
    \multirow{2}{*}{Regression Based} 
                                      & ARX \cite{box2015time}                                   &   regular                        &  qualitative            & single dimensional                      \\ \cline{2-5} 
                                      & IMR \cite{DBLP:journals/pvldb/ZhangS0Y17}                                   &  regular               &   qualitative                       & single dimensional                      \\ \hline
    \multirow{2}{*}{Statistical}      & DPC \cite{DBLP:conf/sigmod/ZhangSW16}                                   &  regular               &  qualitative                        &  single dimensional                     \\ \cline{2-5} 
                                      & IIM \cite{DBLP:conf/icde/ZhangSSW19}                                   &  regular                 &  qualitative                        &  multidimensional                       \\ \hline
    \multirow{2}{*}{MF Based}         & TRMF \cite{DBLP:conf/nips/YuRD16}                                  & regular                 & qualitative                         &  multidimensional                     \\ \cline{2-5} 
                                      & NMF \cite{DBLP:conf/icml/MeiCGH17}                                   & regular                & qualitative                          & multidimensional                      \\ \hline
    \multirow{2}{*}{EM Based}         & EM \cite{DBLP:conf/nips/GhahramaniJ93}                                    &  regular              & qualitative                         & multidimensional                      \\ \cline{2-5}
                                      & EM-GMM \cite{nelwamondo2007missing}                                    &  regular              & qualitative                         & multidimensional                      \\ \hline
    \multirow{2}{*}{MLP Based}        & MLP \cite{DBLP:journals/nca/SharpeS95}                                    &  regular              & qualitative                         & single dimensional                     \\ \cline{2-5} 
                                      & ANN \cite{nordbotten1996neural}                                    &  regular              & qualitative                         & single dimensional                     \\ \hline 
    \multirow{5}{*}{DL Based}         & GRU-D \cite{che2018recurrent}                                 &   regular/irregular              &     qualitative                     & multidimensional                      \\ \cline{2-5} 
                                      & GRUI-GAN \cite{DBLP:conf/nips/LuoCZXY18}                              & regular/irregular                 &  qualitative                        & multidimensional                       \\ \cline{2-5} 
                                      & BRITS \cite{DBLP:conf/nips/CaoWLZLL18}                                 & regular/irregular               &    qualitative                      & multidimensional                      \\ \cline{2-5} 
                                      & E2GAN \cite{DBLP:conf/ijcai/Luo0CY19}                                 & regular/irregular               &  qualitative                         & multidimensional                       \\ \cline{2-5} 
                                      & NAOMI \cite{DBLP:conf/nips/LiuYZZY19}                                 & regular/irregular                  &   qualitative                       & multidimensional                       \\ \hline
    \end{tabular}
  }
\end{table*}

\section{Categorization}
\label{sect:classification}

In this section, we will give a brief introduction of the major approaches to time series imputation. Moreover, we will classify existing time series imputation methods according to the 
principles and techniques they rely on. 

In order to impute the missing values, researchers have proposed many imputation methods to handle the missing values in time series. In this paper,
we mainly conclude 8 kinds of the missing value imputation methods including \textbf{deletion methods}, \textbf{neighbor based methods}, \textbf{constraint based methods},
\textbf{regression based methods}, \textbf{statistical based methods}, \textbf{MF based methods}, \textbf{EM based mathods}, \textbf{MLP based mathods} and \textbf{DL based methods}.
Table~\ref{tab:comparison-of-methods} shows the comparison of these methods we conclude. We will introduce each kind of method respectively as follows.

\textbf{Deletion methods} take a simple strategy that they directly erase the observations that contain missing values from the raw data \cite{mcknight2007missing, wothke2000longitudinal}. 
It is also a commonly adopted strategy when the missing value is not 
high and the deletion of the missing values will not influence the downstream applications. However, when the missing rate reaches some level (in \cite{graham2009missing}, it is 5\%), 
ignoring the missing values and deleting them make the data incomplete and not suitable for downstream applications.

\textbf{Neighbor based methods} \cite{
batista2002study,DBLP:conf/kdd/SongLZ15} find out the imputation value from neighbors, e.g., identified by clustering methods like KNN or DBSCAN.
They first find the nearest neighbors of the missing values through other attributes, and then update the missing values with the mean value of these neighbors. 
Moreover, considering the local similarity, some methods take the last observed valid value to replace 
the blank \cite{amiri2016missing}. 
SRKN (Swapping Repair with K Neighbors) \cite{icde19} in our preliminary study could also be adapted to impute the missing values that are misplaced in other dimensions. 

\textbf{Constraint based methods} \cite{DBLP:journals/pvldb/SongZC015,DBLP:journals/tkde/SongSZCW20} discover the rules in dataset, and take advantage of these rules to impute. 
To apply to time series data, similarity rules such as differential dependencies \cite{DBLP:journals/tods/Song011,DBLP:journals/tkde/Song0C14} or comparable dependencies \cite{DBLP:conf/icde/SongCY11,DBLP:journals/vldb/Song0Y13} could be employed that study the distances or similarities of timestamps as well as values \cite{DBLP:journals/isci/SongZ014}. 
More advanced constraints could be specified in a graph structure \cite{DBLP:conf/icde/0001SLZP15,DBLP:conf/sigmod/ZhuSL0Z14}, such as Petri net, and employed to impute the qualitative values of events in time series \cite{DBLP:journals/pvldb/0001SZL13,DBLP:journals/tkde/0001SZLS16}.
These methods are effective when the data is highly continuous or satisfies certain patterns. For example, when the data is increasing linearly, it is effective and efficient
to take simple methods or clustering methods. And when the rules or constraints are satisfied, constraints based methods outperform others in both time and accuracy \cite{DBLP:conf/sigmod/SongZWY15}.
However, multivariable time series in the real world are not usually satisfied with such rules, thus more general methods are required and learning based methods are 
researched to impute the time series automatically.

\textbf{Regression based methods} 
LOESS~\cite{Cleveland1996} learns a regression model from nearest neighbors for predicting the missing value referring to the complete attributes.
For time series data, autoregressive (AR) models (e.g., ARX \cite{box2015time} and ARIMA \cite{DBLP:journals/ijon/Zhang03}) try to predict missing values from historical data.
More advanced IMR (iterative minimum repairing \cite{DBLP:journals/pvldb/ZhangS0Y17}) provides both anomaly detection and data repair for both anomalies and missing values.
These methods mostly benefit from historical data as well as the accuracy of the nearest neighbors. Thus they could be applied when neighbors are reliable and the time series are highly relative.

\textbf{Statistical based methods} rely on statistical models to impute the missing values \cite{little2019statistical}. 
Simple statistical methods just utilize the data in the original data to impute the missing values, such as take the mean value or median value of the 
attribute to impute \cite{acuna2004treatment, kantardzic2011data}. 
\cite{DBLP:conf/sigmod/ZhangSW16} estimates probability values by statistics on speeds as well as the changes. 
Recently, more advanced IIM (Imputation via Individual models) \cite{DBLP:conf/icde/ZhangSSW19} adaptively learns individual models for various number of neighbors.
Unlike regression based methods which based on just historical data, statistical based models are learned from the whole dataset, including historical data and future data. 
Therefore, they may capture more information from raw data. 

\textbf{Matrix Factorization based methods}
The Matrix Factorization (MF) algorithm tries to impute the value with the Matrix Factorization and reconstruction to find the correlations among the data and complete the missing values which is a classical method of 
collaborative filtering \cite{luo2014incremental}. In recent years MF based approaches are introduced into time series imputation fields \cite{DBLP:conf/nips/YuRD16, DBLP:conf/icml/MeiCGH17}. 
In general, MF based approaches decompose the data matrix into 2 low-dimensional matrices in the meantime extracting the features from original data. And then they try to reconstruct the original matrix and 
in this processing, missing values are imputed. 

\textbf{Expectation-Maximization based methods}
Expectation-Maximization (EM) based methods have been successfully applied to missing data imputation problems \cite{nelwamondo2007missing,DBLP:journals/nca/Garcia-LaencinaSF10,DBLP:conf/nips/GhahramaniJ93}. EM based methods follow a two-stage strategy consisting of the E (Expectation) step and the 
M (Maximization) step which iteratively imputes the missing values with the statistical model parameters
and then updates the statistical model parameters to maximize the possibility of the distribution of the filled data.

\textbf{Multi-Layer Perceptron based methods}
Multi-Layer Perceptron (MLP) based methods employee MLP, which is also called fully connected networks. MLP tries to predict missing value by complete values. It can be divided into 3 parts: input layers, hidden layers and output layers.
In this approach, by minimizing the loss function, the perceptron learns a function to impute missing values by input variables. In \cite{DBLP:journals/nca/SharpeS95}, MLP is used to predict missing values in neural network-based diagnostic systems.
And in \cite{nordbotten1996neural}, MLP is employed to impute Population Census.

Recently, \textbf{deep learning based methods} \cite{che2018recurrent, DBLP:conf/nips/LuoCZXY18, DBLP:conf/ijcai/Luo0CY19, DBLP:conf/nips/CaoWLZLL18} 
mainly deploy Recurrent Neural Network (RNN), since RNN is capable of capturing the time information. In these papers, time information is handled separately and 
attached with more importance. To impute the time series, not only RNN is used, they also combine the models like Gated Recurrent Unit (GRU)
\cite{che2018recurrent, DBLP:conf/nips/LuoCZXY18, DBLP:conf/ijcai/Luo0CY19} 
to extract the long-term information, Generative Adversarial Networks (GAN) \cite{DBLP:conf/nips/LuoCZXY18, DBLP:conf/ijcai/Luo0CY19} to generate the imputed values 
and Bidirectional Recurrent Networks to improve the accuracy \cite{DBLP:conf/nips/CaoWLZLL18}. 

According to the above classification, due to the length, the methods for time series imputation are too many to give a detailed introduction. 
Since among these methods, deep learning based ones are the latest and most powerful, we will discuss 3 latest deep learning methods 
for time series imputation, find the connections and the differences among them.
  
\section{Preliminary}
In this section, we first give our formalization of the imputation tasks. It is because when introducing the aforesaid deep learning methods, 
they formalize the imputation tasks with different symbols and formulas. And in our research, we review them and explain their methods with uniform definitions.

\begin{definitions}[Multivariable Time Series]
We first denote a timestamp lists $\mathbf{T} = (t_0, t_1,..., t_{n-1})$, and the time series $\mathbf{X} = \{\mathbf{x_{t_0}},\mathbf{x_{t_1}}, ...,\mathbf{x_{t_{n-1}}}\}^T$ 
as a sequence of $n$ observations. The $i$-th observation of $\ \mathbf{X}$ is $\mathbf{x_{t_i}}$, which consists of $d$ attributes $\{x_{t_i}^0, x_{t_i}^1, ..., x_{t_i}^d\}$. 
\end{definitions}

After defining the multivariable time series, we use mask matrix $\mathbf{M}$ to denote the missing values.
\begin{definitions}[Mask Matrix]
Mask Matrix $\mathbf{M}$ represents the missing values in $\mathbf{X}$, i.e., $\mathbf{M} \in \mathbb{R}^{n\times d}$. 
And each element of $\mathbf{M}$ is defined as below
\begin{equation}
\mathbf{M}_{t_i}^{j}=
\left\{
  \begin{array}{ll}
    {0} & {\text { if } x_{t_i}^{j} \text { is not observed, i.e. } x_t^j = \text{None}} \\ 
    {1} & {\text { otherwise }}
  \end{array}\right.
\end{equation}
\end{definitions}

To utilize the time information, the time intervals should be recorded with an extra structure.
Therefore, we introduce the time lag, a matrix to represent the time intervals between two adjacent observed values of $\mathbf{X}$.
\begin{definitions}[Time Lag]\
  We use $\mathbf{\delta} \in \mathbb{R}^{n\times d}$ to record the time lag, and we calculate it in an iterative way as follows.
\begin{equation}
  \mathbf{\delta}_{t_{i}}^{j}=\left\{
  \begin{array}{ll}
  {t_{i}-t_{i-1},} & {\mathbf{M}_{t_{i-1}}^{j}=1} \\ 
  {\mathbf{\delta}_{t_{i-1}}^{j}+t_{i}-t_{i-1},} & {\mathbf{M}_{t_{i-1}}^{j-1}==0 \& i>0} \\ 
  {0,} & {i==0}
\end{array}\right.
\end{equation}
\end{definitions}

\begin{examples}
  We now give an example of the time series $\mathbf{X}$, and corresponding timestamp lists $\mathbf{T}$
  \begin{equation}
  \mathbf{X}=
    \left[\begin{array}{cccc}{1} & {6} & {\text { None }} & {9} \\ 
    {7} & {\text { None }} & {7} & {\text { None }} \\ 
    {9} & {\text { None }} & {\text { None }} & {79}\end{array}\right], 
    \mathbf{T}=\left[\begin{array}{c}{0} \\ 
    {5} \\ 
    {13}\end{array}\right]
  \end{equation}
  And we can thus compute the mask matrix $\mathbf{M}$ and the time lag $\mathbf{\delta}$.
  \begin{equation}
    {\mathbf{M}}=\left[\begin{array}{cccc}{0} & {0} & {1} & {0} \\ {0} & {1} & {0} & {1} \\ {0} & {1} & {1} & {0}\end{array}\right],
    \delta=\left[\begin{array}{cccc}{0} & {0} & {0} & {0} \\ {5} & {5} & {5} & {5} \\ {8} & {13} & {8} & {13}\end{array}\right]
  \end{equation}
\end{examples}

\section{Methods}

In this section, we will first give an overall review of the relationships among the given approaches and comparisons of them and then discuss them individually with details.
The main deep learning methods we researched for time series imputation are GRU-D \cite{che2018recurrent}, GRUI-GAN \cite{DBLP:conf/nips/LuoCZXY18},  E$^2$GAN \cite{DBLP:conf/ijcai/Luo0CY19}, 
BRITS \cite{DBLP:conf/nips/CaoWLZLL18} and NAOMI \cite{DBLP:conf/nips/LiuYZZY19}. All of them are deep learning approaches published recently for time series imputation tasks.
Among these methods, recurrent neural network (RNN) and generative adversarial network (GAN) are main architectures that are adopted. 
The reason is that RNN and its variations (e.g., LSTM, GRU) have been proven powerful in modeling sequence data, while GAN has been successfully applied to generation and imputation tasks.

\begin{figure}[t]
  \centering
  \includegraphics[width=0.7\expwidths]{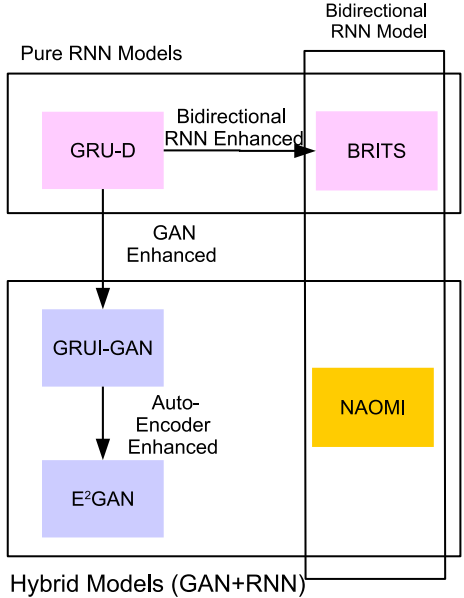}\\
  \caption{The relationships among methods we mainly surveyed.}
  \label{fig:relationships}
\end{figure}

To describe the relationships among these methods, we illustrate the dependencies and common structures of them in Figure~\ref{fig:relationships}.
In Figure~\ref{fig:relationships}, we use arrows to describe the dependencies, for example GRUI-GAN improves the work by using GAN while 
E$^2$GAN is the updated version of GRUI-GAN. And we use boxes to describe the common structures among the methods, for example GRU-D and BRITS are both pure RNN models and 
BRITS and NAOMI both adopt bidirectional RNN structures. This can help us to understand how the time series imputation task is systematically modeled, how the solutions are developed and what progress people 
make in this process. In the following sections,  we will take a progressive order to review them.

\begin{table*}[t]
  \centering
  \caption{Characteristics of the chosen methods}
  \label{tab:characteristics}
    \begin{tabular}{|l|l|l|l|l|l|}
    \hline
    Methodologies & Model Prototype & Specific Models & \begin{tabular}[c]{@{}l@{}}Auto-Encoder \\ Enhanced\end{tabular} & \begin{tabular}[c]{@{}l@{}}Adversarial Training \\ Enhancednced\end{tabular} & \begin{tabular}[c]{@{}l@{}}Bidirectional \\ Enhanced\end{tabular} \\ \hline
    GRU-D         & RNN             & GRU                                                        & --                                                               & --                                                                           & --                                                                \\ \hline
    GRUI-GAN      & Hybrid          & GRU+GAN                                                    & --                                                               & yes                                                                          & --                                                                \\ \hline
    E2GAN         & Hybrid          & GRU+GAN                                                    & yes                                                              & yes                                                                          & --                                                                \\ \hline
    BRITS         & RNN             & Bidirectional RNN                                          & --                                                               & --                                                                           & yes                                                               \\ \hline
    NAOMI         & Hybrid          & RNN+GAN                                                    & --                                                               & yes                                                                          & yes                                                               \\ \hline
    \end{tabular}

\end{table*}

\subsection{Characteristics of Chosen Methods}
\label{sect:characteristics}
In this section, we give the characteristics of the chosen methods in Table~\ref{tab:characteristics} to give a brief introduction and 
a taxonomy of the chosen methods we reviewed. We consider the following criteria:
\begin{itemize}
  \item \emph{Irregular Time Series Awareness}: time series including regular time series with fixed time interval and irregular time series. Both of them are common kinds  
  which are important for classifying the using condition of the methods~\cite{DBLP:conf/sigmod/ZhangSW16, DBLP:conf/sigmod/SongZWY15}. 
  \item \emph{Model Prototype}: model prototype concludes the overall kind of model in the methods, e.g., RNN, GAN and CNN. It is a basic information to classify the model type. If the 
  model prototype is hybrid, it means more than 1 kind of prototype is employed.
  \item \emph{Specific Models}: specific models introduce the specific kinds of model adopted in the methods. The specific models may relate to the basic idea of the methods. 
  \item \emph{Auto-Encoder Enhanced}: auto-encoder structure is an approach that can be applied in the imputation of the data. With the structure of encoder and decoder, 
  it extracts the features from low-dimensional layers and recovery missing values by decoder. Therefore, it can serve as a feature of methods.
  \item \emph{Adversarial Training Enhanced}: adversarial training adopts adversarial structure (e.g., GAN \cite{DBLP:conf/nips/GoodfellowPMXWOCB14} and CGAN \cite{DBLP:journals/corr/MirzaO14})
  to enhance the model. It takes the idea of generative adversarial structure with generator and discriminator. Large amount of models can be enhanced with such idea.
  \item \emph{Bidirectional Enhanced}:  Bidirectional RNN trains 2 models in forward direction and backward direction respectively with RNN and then combines them into the same loss function \cite{DBLP:journals/nn/GravesS05}.
  This idea is vital in data imputation tasks since both previous series and future series of missing values are known. Therefore, bidirectional structure benefits from both backward and forward training processing. Such idea is 
  adopted in \cite{DBLP:conf/nips/LiuYZZY19,DBLP:conf/nips/CaoWLZLL18}.
  
\end{itemize}

\subsection{GRU-D}
\label{subsect:GRUD}
GRU-D is proposed by \cite{che2018recurrent} as one of the early attempts to impute time series with deep learning models.
It is the first one among the 5 researched paper to systematically model missing patterns into RNN for time series 
classification problems. It is also the first research to exploit that, RNN can model multivariable time series with the informativeness
from the time series. 
Former works like \cite{lipton2016directly, choi2016doctor} attempted to impute missing values with RNN by concatenating timestamps and raw data, i.e., regard 
timestamps as one attribute of raw data. But in \cite{che2018recurrent}, the concept \textbf{time lag} is first proposed. 
In this paper, Gated Recurrent Unit (GRU) is first adopted to generate missing values. In each layer of GRU, since the input can contain missing values, 
they replace the input $x_{t_i}^j$ with a combination of the existing values $x_{t_i}^j$ and statistical values, element-wise multiplied with $\mathbf{M}$ 
and $\mathbf{1}-\mathbf{M}$ respectively.
$$
x_{t_i}^{j} \leftarrow m_{t_i}^{j} x_{t_i}^{j}+\left(1-m_{t_i}^{j}\right) \tilde{x}^{j}
$$
where $\tilde{x}$ can be one of the mean value, last observed value or concatenation of $\left[\mathbf{x_i}; \mathbf{m_i};\delta_i\right]$. 

\begin{figure}[t]
  \subfigure[GRU]{
    \label{fig:GRUD1}
    \includegraphics[width=0.45\figwidths]{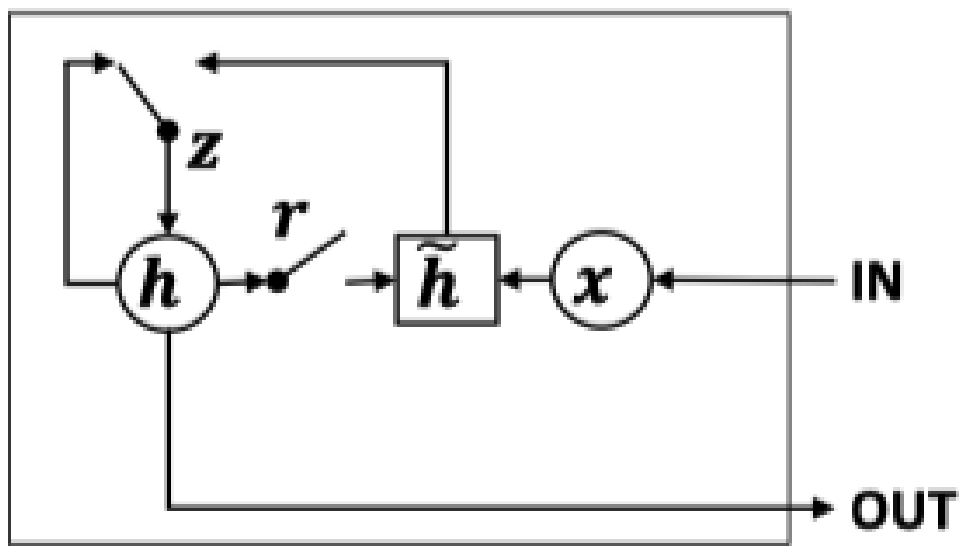}
  }
  \subfigure[GRU-D]{
    \label{fig:GRUD2}
    \includegraphics[width=0.45\figwidths]{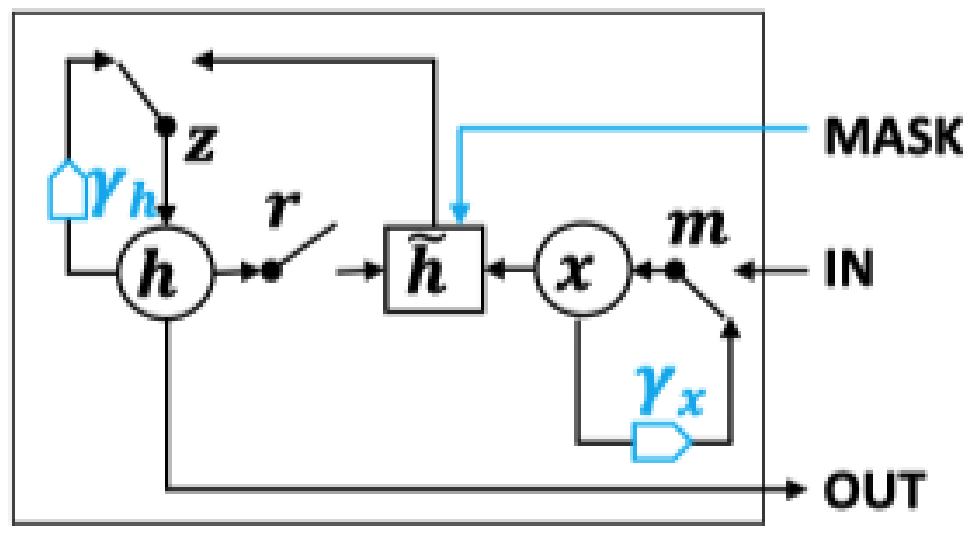}
  }
  \caption{Model of GRU and GRU-D. Images extracted from \cite{che2018recurrent}.}
  \label{fig:GRUD}
\end{figure}

The main contribution of this paper is the GRU based model GRU-D and the proposition of \textbf{decay rate}.
To address the imputation of the missing values, they discover that 
\begin{itemize}
  \item The missing variables tend to be close to some default value if its last observation happens a long time ago. 
  \item The influence of the input variables will fade away over time if the variable has been missing for a while.
  \end{itemize}
And then they propose \textbf{decay rate} $\gamma$, which is defined as below 
$$
\gamma_{t_i} = \exp({-\max{(\mathbf{0},\mathbf{W}_{\gamma}\mathbf{\delta_{t_i}})}})
$$
The decay rate tries to model the impact of the other values have on the missing values. In brief, it guarantees that the larger the time intervals are, 
the less their influence on imputing the missing values. And then they replace the input variable as  
$$
x_{t_i}^{j} \leftarrow m_{t_i}^{j} x_{t_j}^{j}+\left(1-m_{t_i}^{j}\right) \gamma_{\boldsymbol{x}_{t_i}}^{j} x_{t_i^{\prime}}^{j}+\left(1-m_{t_i}^{j}\right)\left(1-\gamma_{\boldsymbol{x}_{t_i}}^{j}\right) \tilde{x}^{j}
$$ 
Therefore, as illustrated in Figure~\ref{fig:GRUD}, the GRU-D model is proposed with 2 different trainable decays $\gamma_{\boldsymbol{x}}$ and $\gamma_{\boldsymbol{h}}$, where 
$\gamma_{\boldsymbol{x}}$ is the input decay rate and the $\gamma_{\boldsymbol{h}}$ is the decay rate for the hidden state.

\subsection{GRUI-GAN}
\label{subsect:GRUI}
\begin{figure}[t]
  \subfigure[GRU]{
    \label{fig:GRUI1}
    \includegraphics[width=0.45\figwidths]{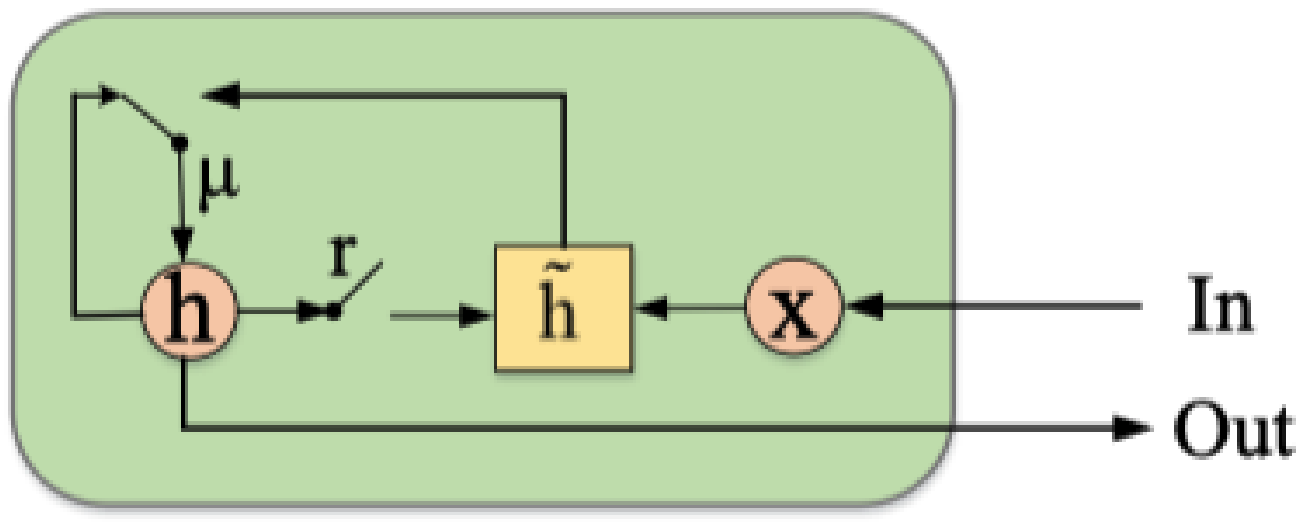}
  }
  \subfigure[GRU-I]{
    \label{fig:GRUI2}
    \includegraphics[width=0.45\figwidths]{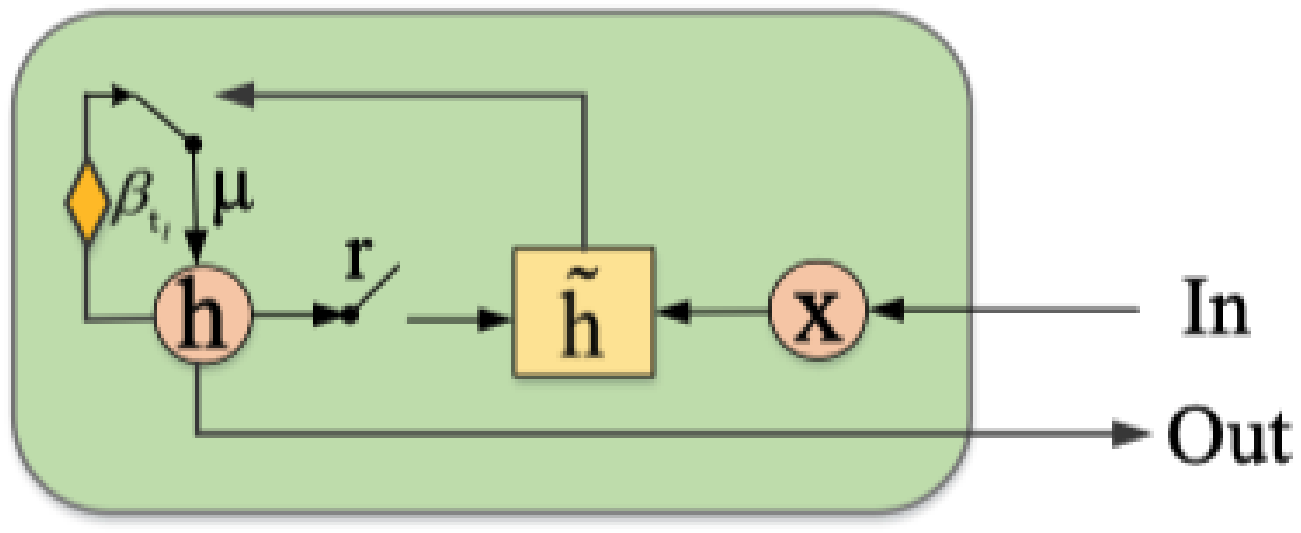}
  }
  \caption{Model of GRU and GRU-I. Images extracted from \cite{DBLP:conf/nips/LuoCZXY18}.}
  \label{fig:GRUI}
\end{figure}

In \cite{DBLP:conf/nips/LuoCZXY18}, GRU-I is proposed as the recurrent unit to capture the time information. As Figure~\ref{fig:GRUI} illustrates, 
it follows the structure of GRU-D in Section~\ref{subsect:GRUD} with the removal of the input decay. Therefore, there is no innovation in the 
RNN part as well as the decay rate.

\begin{figure*}[t]
  \centering
  \includegraphics[width=1.2\expwidths]{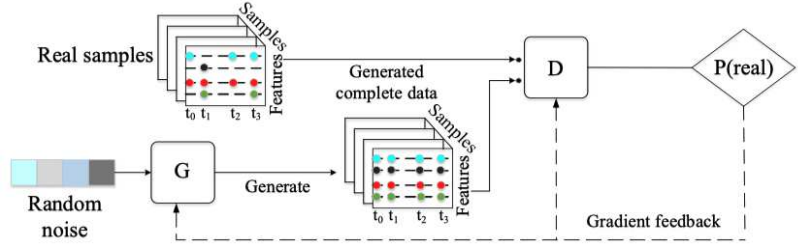}\\
  \caption{The structure of the GRUI-GAN. Image extracted from \cite{DBLP:conf/nips/LuoCZXY18}.}
  \label{fig:GRUI-GAN}
\end{figure*}

The main contribution of this paper locates in the GAN structure. Figure~\ref{fig:GRUI-GAN} shows the structure. 
The Generative Adversarial Network (GAN) structure is made up of a generator (G) and a discriminator (D). 
The G learns a mapping $G(z)$ that tries to map the random noise vector $z$ to realistic time series. 
The D tries to find a mapping $D(.)$ that tells us the input data's probability of being real.
Therefore, in this paper, the model takes a random noise as the input of the GAN model, which means the generating is a 
random process. Both G and D are based on GRU-I, and it takes lots of time to train the model to get the data imputed.

The GRUI-GAN takes advantage of the ability of GAN in imputation, which has been proven powerful in image imputation such as \cite{pathak2016context}.
And the adversarial structure improves accuracy. Moreover, the paper adopts a WGAN structure, which improves the stability of the 
learning stage, get out of the problem of mode collapse and makes it easy for the optimization of the GAN model.

However, this model is not practical since the accuracy of the generative model seems not stable with a random noise input. And it also 
makes the model hard to converge.

\begin{figure}[t]
  \centering
  \includegraphics[width=\expwidths]{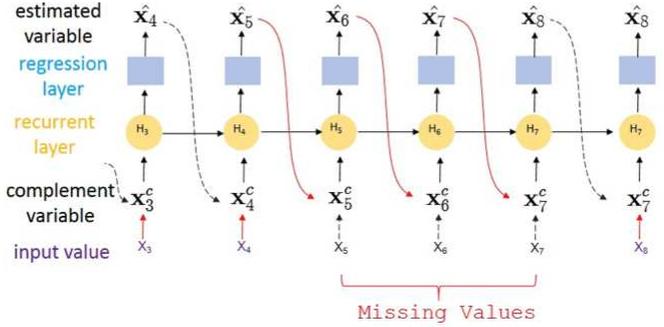}\\
  \caption{The structure of the BRITS. Image extracted from \cite{DBLP:conf/nips/CaoWLZLL18}.}
  \label{fig:BRITS}
\end{figure}

\subsection{BRITS}
\label{sect:BRITS}
Unlike former methods, BRITS \cite{DBLP:conf/nips/CaoWLZLL18} is totally based on RNN structure and proposes imputation with unidirectional dynamics. 
Time lag (corresponding to "time gaps" in \cite{DBLP:conf/nips/CaoWLZLL18}) is also employed since the time series may be irregular.
Similar to the idea of decay rate $\gamma$ from GRU-D introduced in Section~\ref{subsect:GRUD}, they propose \textbf{temporal decay factor} 
$\gamma_t = \exp{(-max\left(0,\mathbf{W}_{\gamma}\delta_t + \mathbf{b}_{\gamma}\right))}$. Compared to GRU-D where the time lags are considered in input and serve as 
the decay rate, in BRITS the hidden states update with the decay rate $\gamma$.
It means when updating the hidden state, the old hidden state decays according to the time duration recorded in the time lags. Hence, the model is updated by:
\begin{equation}
\begin{aligned} 
\hat{\mathbf{x}}_{t} &=\mathbf{W}_{x} \mathbf{h}_{t-1}+\mathbf{b}_{x} \\ 
\mathbf{x}_{t}^{c} &=\mathbf{m}_{t} \odot \mathbf{x}_{t}+\left(1-\mathbf{m}_{t}\right) \odot \hat{\mathbf{x}}_{t} \\
 \gamma_{t} &=\exp \left\{-\max \left(0, \mathbf{W}_{\gamma} \delta_{t}+\mathbf{b}_{\gamma}\right)\right\} \\
 \mathbf{h}_{t} &=\sigma\left(\mathbf{W}_{h}\left[\mathbf{h}_{t-1} \odot \gamma_{t}\right]+\mathbf{U}_{h}\left[\mathbf{x}_{t}^{c} \circ \mathbf{m}_{t}\right]+\mathbf{b}_{h}\right) \\ 
\ell_{t} &=\left\langle\mathbf{m}_{t}, \mathcal{L}_{e}\left(\mathbf{x}_{t}, \hat{\mathbf{x}}_{t}\right)\right\rangle 
\end{aligned}
\end{equation}

The former model named RITS is the unidirectional version of the proposed methods in \cite{DBLP:conf/nips/CaoWLZLL18}. 
As the bidirectional version, BRITS employs bidirectional RNN by utilizing the bidirectional recurrent dynamics, i.e., they train 2 models in forward direction and 
backward direction respectively \cite{DBLP:journals/nn/GravesS05}. 
Thus consistency loss is introduced to take the losses of both directions into consideration.

To conclude, in BRITS, time lags are still adopted to deal with irregular time series. Only RNN is used to model the time series. We can also conclude from the model and the experiments that 
bidirectional RNN contributes to a higher performance since the unidirectional model may suffer from bias exploding problem \cite{DBLP:conf/nips/BengioVJS15}.

\begin{figure*}[t]
  \centering
  \includegraphics[width=1.1\expwidths]{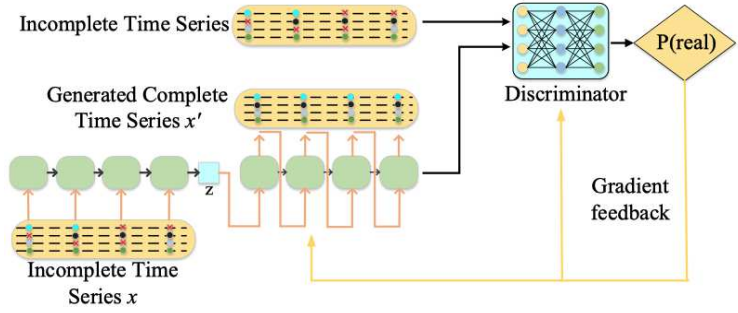}\\
  \caption{The structure of the E$^2$GAN. Image extracted from \cite{DBLP:conf/ijcai/Luo0CY19}.}
  \label{fig:E2GAN}
\end{figure*}    
\subsection{E$^2$GAN}
E$^2$GAN \cite{DBLP:conf/ijcai/Luo0CY19} is another work based on GAN. While the GRUI-GAN in Section~\ref{subsect:GRUI} takes a random noise vector 
as input, which takes lots of time to train, E$^2$GAN adopts an auto-encoder structure based on GRUI to form the generator. The overall structure of their model 
is in Figure~\ref{fig:E2GAN}.

In E$^2$GAN, concepts including mask, time lag, decay rate and GRUI are all reserved without improvement, thus there is no innovation in the GRUI structure.
The main contribution is the auto-encoder structure they adopt in the generator. This is a common strategy taken by image generation and imputation 
such as Context-Encoder \cite{pathak2016context}, PixelGANs \cite{isola2017image}, but not a common strategy in RNN based GAN. 
Since the input of the model is the original time series, the model compresses the input incomplete time series $\mathbf{X}$ into a low-dimensional vector $z$
with the help of the GRUI. And then the reconstructing part will reconstruct the complete time series $\mathbf{X'}$ to fool the discriminator. 
And the discriminator of the method attempts to distinguish actual incomplete time series $\mathbf{X}$ and
the fake but complete sample $\mathbf{X'}$ through the adoption of recursive neural network. The framework of the discriminator is also an encoder.

E$^2$GAN takes an encoder-decoder RNN based structure as the generator, which tackles the difficulty of training the model and the accuracy. 
So far, according to the experiments in the paper, E$^2$GAN has achieved state-of-the-art and outperforms other existing methods.

\subsection{NAOMI}
\begin{figure}[t]
  \centering
  \includegraphics[width=\expwidths]{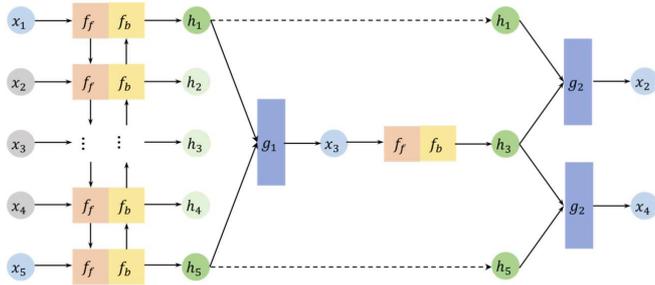}\\
  \caption{The structure of the NAOMI. Image extracted from \cite{DBLP:conf/nips/LiuYZZY19}.}
  \label{fig:naomi}
\end{figure}

NAOMI (\textbf{N}on-\textbf{A}ut\textbf{O}regressive \textbf{M}ultiresolution \textbf{I}mputation \cite{DBLP:conf/nips/LiuYZZY19})
proposes a non-autoregressive model which conditions both previous values but also future values, i.e., equipped with bidirectional 
RNN like BRITS introduced in Section~\ref{sect:BRITS}. Since in the imputation tasks, future values and historical values are both observed,
the intuition is to take advantage of both values and train bidirectional models for them. As illustrated in Figure~\ref{fig:naomi}, $f_f$ and $f_b$ are 
forward and backward RNN respectively, thus the hidden state $h_t$ is a joint hidden state concatenated by $h^f_t$ and $h^b_t$.

Moreover, a special predicting strategy is performed in this paper. They adopt a \emph{divide and conquer strategy}. As it is shown in Figure~\ref{fig:naomi}, 
with 2 known values $x_1$ and $x_5$, they first predict the midpoint $x_3$ by $x_1$ and $x_5$ with proposed bidirectional RNN models, and then $x_3$ is updated and 
utilized to predict $x_2$ and $x_4$ respectively. Thus a fine-grained prediction is performed.
Finally, adversarial training is taken to enhance the model. 

However, in NAOMI, time gaps are ignored and the data is injected into the RNN model without timestamps. It suggests the model is not aware of 
irregular time series although we can still take them as input by removing their timestamps directly.

\section{Conclusion}
In this paper, we give a brief introduction to the imputation methods for time series. We propose that existing methods can be classified into 3 main classed: 
deletion methods, traditional methods, and learning based methods. And we introduce our classification in detail. Moreover, we investigate existing deep learning methods
for time series imputation, since they outperform others and make great progress recently. We mainly researched 3 deep learning methods including 
GRU-D, GRUI-GAN, and E$^2$GAN. All of them based on RNN, and the latter two also adopt GAN for more accurate imputation. We also find the relationships among them: GRUI-GAN 
is based on the definitions from GRU-D, and E$^2$GAN improves the generator of the GRUI-GAN with auto-encoder. And so far, E$^2$GAN achieves state-of-the-art. 

Since the imputation problem is fundamental, we believe with these methods, the filled data would benefit downstream applications in many aspects. And as we observed, most 
of the techniques in other fields can be adopted in this task since time series data is everywhere. In the future, we would like to see the time information can be utilized 
properly, and the methods can be more general and accurate so that we would not need to choose the best one from too many methods, 
and the missing data of the time series would not be a problem.

\section{Future Research Opportunities}
Based on our observation from surveying the development of time series imputation methodologies, we try to highlight some potential research opportunities in this field.
Most existing researches mainly focus on the structure of RNN and try to use bidirectional RNN, Auto-Encoder structure and GAN to enhance the model. With the rapid development in the 
deep learning society (especially Natural Language Processing (NLP) where time series are also highly concerned),  some techniques have reached better performance (e.g., attention models). 
These models can be considered to enhance the imputation models.
Further, most existing methods ignore the missing of timestamps which can also appear obscurely \cite{DBLP:journals/pvldb/SongC016}. 
Therefore, there is still demand for such techniques. Existing methods can be extended to impute missing timestamps.
Moreover, query answering without directly imputing missing values is another perspective of dealing with missing values. 
Under such scenarios, specific values do not need imputation, and consistent queries in inconsistent probabilistic databases should be generated.

\subsection{Attention Mechanism Enhanced}
In recent years, the attention mechanism has been shown successful in deep learning society, especially in NLP fields.
When adopted in RNN, the attention mechanism allocates weights for each hidden state to draw information from the sequence.
With such mechanism, the model is improved to capture latent patterns in historical data, thus may benefit time series imputation.
Compared to existing RNN models (e.g., LSTM and GRU) which already take long-term dependencies into consideration, the attention mechanism for instance temporal attention enables the model to 
see features and status globally. However, LSTM and GRU will still lose long-term information due to the forget gate unit.

Recently, pure attention models are proposed without RNN. The Transformer proposed in \cite{DBLP:conf/nips/VaswaniSPUJGKP17} is one of the popular frameworks.
In the proposed Transformer framework, it only adopts an attention layer called self-attention, which is computed as:
$$
\operatorname{Attention}(Q, K, V)=\operatorname{softmax}\left(\frac{Q K^{T}}{\sqrt{d_{k}}}\right) V
$$
where $Q, K, V$ are queries, keys and values respectively, and $d_k$ is the dimension of the input.

Accepting a single sequence as input, the self-attention mechanism relates different positions of the input and tries to compute a representation of the sequence.
Without applying RNN, the Transformer relies entirely on the self-attention layers to former an encoder-decoder structure, which is similar to the auto-encoder introduced in 
Section~\ref{sect:characteristics}. Such a structure provides the ability to extract high-dimensional features for reconstructing, which benefits tasks like machine translation introduced in \cite{DBLP:conf/nips/VaswaniSPUJGKP17}.

For improving the performance of data imputation, due to the effectiveness of the attention mechanisms, models based on attention mechanisms may also address the time series imputation problems.
And two aforementioned categories of the attention mechanisms including temporal attention and self-attention are both potential techniques which may benefit the time series imputation.
Moreover, with the idea of removing RNN and leveraging only attention mechanisms, structures like the Transformer may contribute to a new framework for the imputation tasks.

To summary, two categories of attention mechanisms including temporal attention and self-attention may bring future opportunities on time series imputation. And the pure attention frameworks are also 
new directions to model time series.

\subsection{Imputing Missing Timestamps}
Missing timestamps often appear obscurely \cite{DBLP:journals/pvldb/SongC016}, e.g., denoted by \texttt{00:00:00}. Most of existing methods mainly focus on the missing values of the time series.
However, once timestamps are missing, these methods may fail to capture the information of time and unable to obtain accurate imputation results. 
Thus, an extension of existing methods to impute missing timestamps is potentially appropriate direction to deal with such scenarios. 

\subsection{Consistent Query Answering}
Following \cite{DBLP:conf/sigmod/LianCS10}, query answering without determining the specific imputation of each missing value is crucial in probabilistic databases \cite{DBLP:journals/vldb/DalviS07}, 
when data from many sources can be inconsistent and uncertain. Therefore, consistent query answering (CQA) is needed. Missing values data in CQA problem increase 
the difficulty of answering the query consistently. Both the inconsistent data from different sources and missing values should be considered. Therefore, a combination of data imputation methods and 
CQA methods can be a potential approach.

\bibliographystyle{abbrv}
\bibliography{survey-imputation}

\end{document}